\documentclass[runningheads]{llncs}
\usepackage{graphicx}
\usepackage{comment}
\usepackage{float}
\usepackage{amsmath,amssymb} 
\usepackage{color}
\usepackage{multirow}
\usepackage{booktabs}
\usepackage{array}

\makeatletter
\def\thickhline{%
	\noalign{\ifnum0=`}\fi\hrule \@height \thickarrayrulewidth \futurelet
	\reserved@a\@xthickhline}
\def\@xthickhline{\ifx\reserved@a\thickhline
	\vskip\doublerulesep
	\vskip-\thickarrayrulewidth
	\fi
	\ifnum0=`{\fi}}
\makeatother

\newlength{\thickarrayrulewidth}
\begin{document}
\pagestyle{headings}
\mainmatter

\title{MVP: Multimodality-guided Visual Pre-training} 

\titlerunning{ECCV-22 submission ID \ECCVSubNumber} 
\authorrunning{ECCV-22 submission ID \ECCVSubNumber} 
\author{Anonymous ECCV submission}
\institute{Paper ID \ECCVSubNumber}

\titlerunning{MVP: Multimodality-guided Visual Pre-training}

\author{\!Longhui Wei\textsuperscript{1,2}, Lingxi Xie\textsuperscript{2}, Wengang Zhou\textsuperscript{1}, Houqiang Li\textsuperscript{1}, Qi Tian\textsuperscript{2}\!}

\authorrunning{L. Wei, L. Xie, et al.}
%
\institute{\textsuperscript{1}University of Science and Technology of China,\quad\textsuperscript{2}Huawei Inc.,\\
\email{weilh2568@gmail.com}, \email{198808xc@gmail.com},
\email{zhwg@ustc.edu.cn}\\
\email{lihq@ustc.edu.cn},
\email{tian.qi1@huawei.com}}
\maketitle

\begin{abstract}
{Recently, masked image modeling (MIM) has become a promising direction for visual pre-training. In the context of vision transformers, MIM learns effective visual representation by aligning the token-level features with a pre-defined space (\textit{e.g.}, BEIT used a d-VAE trained on a large image corpus as the tokenizer). In this paper, we go one step further by introducing guidance from other modalities and validating that such additional knowledge leads to impressive gains for visual pre-training. The proposed approach is named Multimodality-guided Visual Pre-training (MVP), in which we replace the tokenizer with the vision branch of CLIP, a vision-language model pre-trained on 400 million image-text pairs. We demonstrate the effectiveness of MVP by performing standard experiments, \textit{i.e.}, pre-training the ViT models on ImageNet and fine-tuning them on a series of downstream visual recognition tasks. In particular, pre-training ViT-Base/16 for $300$ epochs, MVP reports a 52.4\% mIoU on ADE20K, surpassing BEIT (the baseline and previous state-of-the-art) with an impressive margin of 6.8\%.}
\end{abstract}

\section{Introduction}
\label{sec:intro}
Deep neural networks have been a fundamental tool for computer vision, yet they often require a large amount of labeled training data~\cite{deng2009imagenet} and the model can sometimes bias towards the semantic labels. A promising direction to alleviate the issues is unsupervised visual pre-training, which has been attracting increasing attentions in both academia and industry. After the early efforts based on geometries~\cite{noroozi2016unsupervised,wei2019iterative} and image generation~\cite{pathak2016context,vincent2008extracting}, the emerge of contrastive learning~\cite{he2020momentum,chen2020improved,chen2020simple,tian2020makes,tian2019contrastive,wei2021exploring} has made a great progress in learning from large-scale image data. Without semantic annotations, these approaches report competitive downstream transfer performance, sometimes even surpassing the supervised counterpart~\cite{he2019rethinking}.

Recently, vision transformers~\cite{dosovitskiy2020image,touvron2021training} have been validated effective in a wide range of visual recognition tasks, but these models are also shown to heavily rely on large-scale training data. To alleviate the burden, researchers start investigating unsupervised pre-training. Besides applying contrastive learning~\cite{MOCOV3,DINO}, another interesting methodology is named masked image modeling (MIM)\footnote{It borrows the framework of masked language modeling (MLM)~\cite{lan2019albert,devlin2018bert} from natural language processing.}. MIM~\cite{BEIT,MAE,MaskFeat,SimMIM,PeCo,ibot} removes part of image patches from input and requires the target model to recover the missing contents. Currently, there are two main directions for MIM: one~\cite{BEIT,PeCo} is to predict the tokenized features (\textit{e.g.}, by d-VAE~\cite{dvae} or VQ-VAE~\cite{VQVAE}), and the other~\cite{MAE,SimMIM} is to predict pixel-level information. MIM works particularly well for vision transformer models, \textit{e.g.}, when the pre-trained backbone is allowed to be fine-tuned, state-of-the-art image classification accuracy is reported on ImageNet-1K~\cite{deng2009imagenet}. But, we note that such models are weak when the backbone is frozen -- for example, BEIT~\cite{BEIT} reports a $37.6\%$ accuracy in the linear probing test on ImageNet-1K; MAE~\cite{MAE} improves it to $67.8\%$, but it is still significantly lower than that reported by contrastive learning (\textit{e.g.}, DINO reports $78.2\%$). This makes us conjecture that the pre-trained model learns relatively weak semantic features for visual representation.

\begin{figure}[!t]
\centering
\includegraphics[width=11cm]{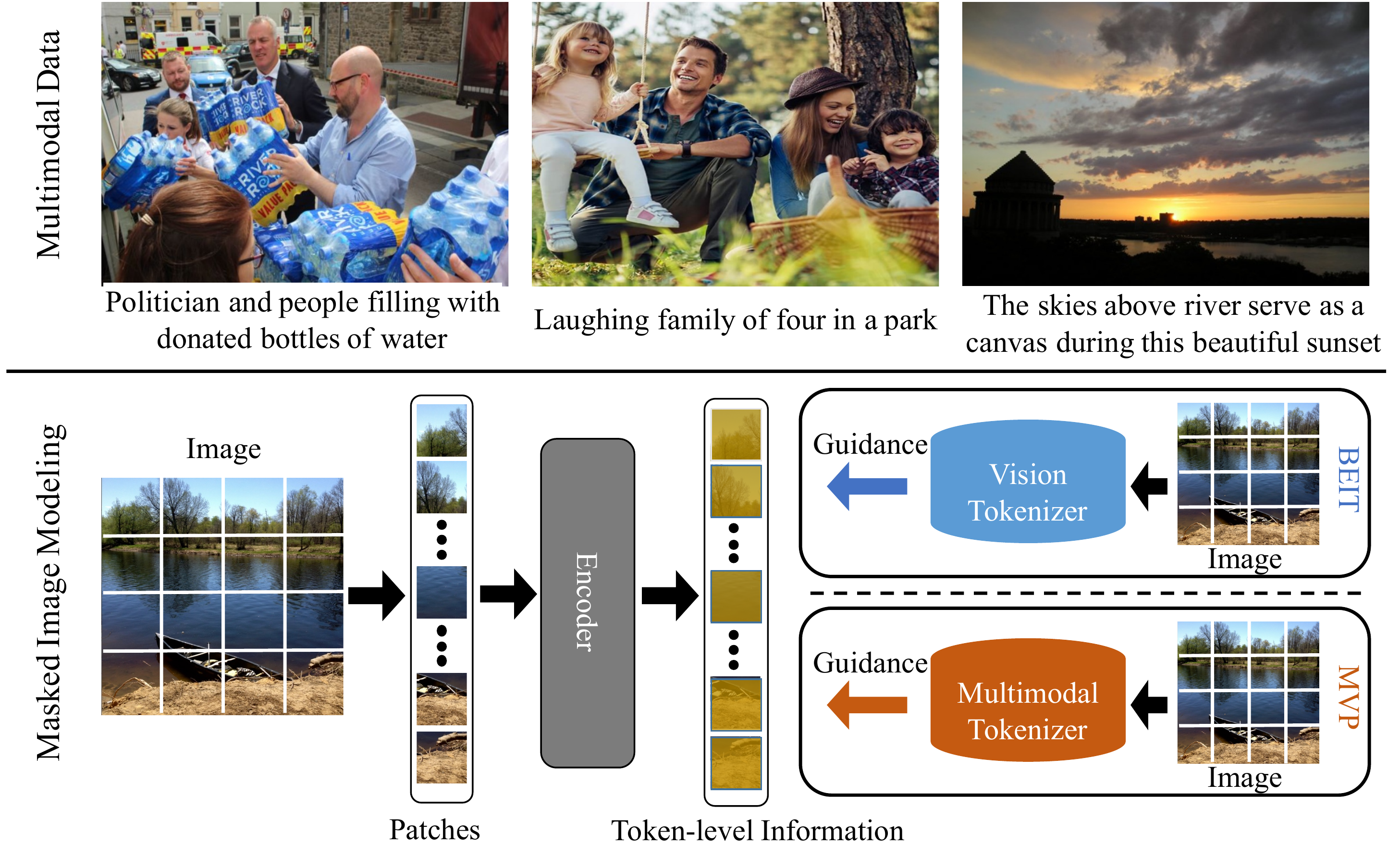}
\caption{The motivation of our MVP. The image-text pairs of CC3M~\cite{cc3m} are shown at the top. For each case, the caption can depict some important semantic contents. The pipeline of masked image modeling is presented at the bottom, and our MVP simply replaces the vision tokenizer in BEIT with the multimodal tokenizer.}
\label{fig:introduction}
\end{figure}

The goal of this paper is to enhance the semantics for MIM. For this purpose, we present Multimodality-guided Visual Pre-training (MVP), a single yet effective framework that incorporates multimodal information into MIM, in particular, the BEIT framework~\cite{BEIT}. As shown in Figure~\ref{sec:intro}, our motivation is simple that multimodal data can provide more semantic knowledge. Therefore, instead of using a tokenizer that was pre-trained with pure image data, we replace it with a tokenizer that is pre-trained with image-text pairs. We expect the latter to provide weak semantic guidance (since the tokenizer is required to align vision and language) and open-domain representation ability (the texts are not constrained by a set of pre-defined classes). To the best of our knowledge, this is the first work that investigates the use of multimodal pre-training on the MIM framework.

MVP is easily implemented upon BEIT, \textit{i.e.}, directly changing the tokenizer. In particular, we refer to the pre-trained model of CLIP~\cite{2021Learning} that has seen 400 million image-text pairs, and directly take the vision branch as the tokenizer. It replaces the original tokenizer pre-trained by d-VAE~\cite{dvae}. Other parts of BEIT are nearly unchanged expect for the prediction pretext task. Interestingly, such a simple modification brings large benefits on a series of downstream tasks. MVP reports a $75.4\%$ accuracy on ImageNet-1K linear probing, which significantly surpasses the numbers of BEIT ($37.6\%$) and MAE ($67.8\%$), demonstrating its strong ability of semantic learning. In the fine-tuning test, MVP reports $84.4\%$ and $86.3\%$ accuracy with ViT-Base/16 and ViT-Large/16 backbones, respectively, both of which surpass the BEIT baseline by more than $1\%$. Most notably, when the pre-trained backbone is transferred for semantic segmentation on ADE20K~\cite{ADE20K}, MVP with a ViT-Base/16 backbone achieves a $52.4\%$ mIOU, which outperforms all existing MIM-based methods by a remarkable margin of $3.6\%$.

The main contributions of this paper can be summarized as follows:
\begin{itemize}
\item
We analyze the recent masked image modeling (MIM) based pre-training methods lack of semantics knowledge, and then firstly point out they can be enhanced with the guidance of other modalities.
\item
We design a simple yet effective algorithm to improve the transfer performance of MIM-based visual pre-training. By resorting to a tokenizer pre-trained with multimodal data (image-text pairs), MVP learns richer semantic knowledge for each image.
\item
We evaluate the effectiveness of MVP with extensive experiments, and the results clearly demonstrate the advantages of MVP over the recently proposed visual pre-training methods. 
\end{itemize}

\section{Related Work}

In the deep learning era, a fundamental methodology for visual recognition is to train deep neural networks. In the scenarios with insufficient labeled training data, a popular pipeline is to pre-train the model with labeled/unlabeled data from other sources (\textit{e.g.}, ImageNet~\cite{deng2009imagenet}) and transfer the model to specific domains. This paper focuses on unsupervised (self-supervised) pre-training. In this section, we review two sub-topics in this field, namely, visual pre-training and multimodal pre-training.

\subsection{Visual Pre-training}
Currently, as self-supervised learning methods with contrastive loss~\cite{he2020momentum,chen2020improved,tian2019contrastive,tian2020makes} heavily boost the transfer performance of visual models, self-supervised learning has become the mainstream in the visual pre-training field. For example, He~\emph{et al.}~\cite{he2020momentum} proposed a momentum contrast framework to reduce the limitation of requirement on batch size, and significantly pushed forward the transfer performance of pre-training models. Concurrently, SimCLR~\cite{chen2020simple} has been proposed to verify the effect of different data augmentation strategies in contrastive learning based methods. Moreover, target to avoid the confusion of sampled negative noise, BYOL~\cite{grill2020bootstrap} was designed to achieve competitive results by simply pushing the positive pairs together. Recently, vision transformer based architectures~\cite{VIT,DEIT} have been validated in a wide range of visual tasks  compared with traditional convolutional networks. To improve the transfer performance of vision transformers, some self-supervised works~\cite{DINO,MOCOV3} were further proposed for effectively pre-training the vision transformer backbones. 

As masked language modeling (MLM) based methods~\cite{devlin2018bert,lan2019albert} achieve great success in the natural language processing field, more and more researchers expect to design similar pretext tasks to enhance the visual pre-training models. Motivated by this, masked image modeling (MIM) based approaches have been designed recently and achieved competitive results. For example, by simply designing a pretext task with the visual token prediction of each masked image patch, BEIT~\cite{BEIT} heavily enhanced the transfer performance of visual models. Moreover, with the pixel-level information reconstruction of each masked patch, MAE~\cite{MAE} further improved the final results. Concurrently, some similar MIM-based schemes~\cite{SimMIM,MaskFeat,CAE} have been proposed and pushed forward the development of visual pre-training. In this work, we also utilize the MIM-based framework but design a special multimodality-driven pretext task to guide the visual models learning more multimodal semantic knowledge.

\subsection{Multimodal Pre-training}
Information is commonly reserved as different modalities in real scenarios. Because of its increasing importance, multimodal pre-training has been attracting more and more researchers~\cite{2021Learning,Align,Visualbert,Uniter,Oscar}. The existing multimodal pre-training works can be mainly summarized with two mainstream directions according to the network architecture,~\emph{i.e.}, one-stream multimodal network architecture based methods and two-stream multimodal network architecture based methods. For most works with one-stream multimodal architecture~\cite{Oscar,Uniter,Visualbert}, they usually encoded the language and image into discrete tokens, and then fused the information of these tokens in the early stage of network. Though these works can perform well on fusing different modalities, their inference efficiencies are relatively poor. To address this challenge, researchers tend to utilize two-stream network architecture~\cite{2021Learning,Align} for processing the interaction of different modalities. For example, CLIP~\cite{2021Learning}, a recent state-of-the-art multimodal pre-training approach, extracted each modality information with one alone branch, and then simply aligned the extracted feature from each branch into a common multimodal representation space. Benefiting from it pre-trained on $400$ million image-text pairs, CLIP heavily pushed forward the transfer performance of multimodal models on a wide range of downstream tasks. In this work, we have no expects to design a new multimodal pre-training framework, but utilize a pre-trained multimodal model to guide the semantic knowledge learning of visual pre-training models.

\section{Our Approach}
\label{sec:method}

\subsection{Problem Setting}
Given a large-scale image dataset $\mathcal{D}=\{\mathbf{I}_n\}_{n=1}^N$, the goal of visual pre-training is to guide a computational model to learn transferable knowledge on $\mathcal{D}$. After pre-training, the visual backbone with pre-trained parameters will be transferred into different visual downstream tasks for improving the corresponding performances. Currently, many self-supervised visual pre-training schemes are proposed to enhance the transfer performance of models by designing different pretext tasks. Given a pretext task as $\mathcal{L}(\cdot)$ and visual model as $f^\mathrm{vis}(\mathbf{I}_n;\boldsymbol{\theta})$, the goal of self-supervised visual pre-training can be written as:
\begin{eqnarray}
\label{eqn:problem_seeting}
{\min_{\boldsymbol{\theta}}\mathbb{E}_{\mathbf{I}_n\in\mathcal{D}}[\mathcal{L}(\mathbf{I}_n,f^\mathrm{vis}(\cdot;\boldsymbol{\theta}))]},
\end{eqnarray}
where $\mathbb{E}_{\mathbf{I}_n\in\mathcal{D}}[\cdot]$ denotes the expectation on the entire training dataset, $\mathcal{D}$.

\subsection{Masked Image Modeling with Tokenizer}
As presented in Eqn (\ref{eqn:problem_seeting}), the core of self-supervised visual pre-training is to design a proper pretext task. As vision transformers~\cite{VIT,DINO} achieve competitive results on visual recognition tasks, researchers~\cite{BEIT,PeCo} resort to BERT-style schemes in natural language processing and design masked image modeling (MIM) based pretext tasks to guide the visual pre-training. Given an image $\mathbf{I}\in\mathcal{R}^{H\times{W}\times3}$, it can be divided into several image patches $\{\mathbf{p}_1,\mathbf{p}_2,\ldots,\mathbf{p}_M\}$, where $M$ represents the number of patches, and these patches are further encoded as isolated tokens $\{\mathbf{t}_1,\mathbf{t}_2,\ldots,\mathbf{t}_M\}$ in the vision transformer backbone. MIM-based methods usually mask percentages of tokens as $\{\mathbf{t}_1,\ldots,\hat{\mathbf{t}}_m,\ldots,\mathbf{t}_M\}$, where $\hat{\mathbf{t}}_m$ represents that the $m$-th token is replaced by a \textsf{MASK} token. Then, MIM-based methods~\cite{PeCo,BEIT,MaskFeat} utilize the pretext by predicting the information (provided by the Tokenizer) of these masked tokens for pre-training visual models. 

Take BEIT (a current state-of-the-art MIM-based method) as an example, the optimization goal of MIM with Tokenizer can be formulated as:
 \begin{eqnarray}
\label{eqn:BEIT}
{\mathcal{L}_\mathrm{BEIT}}&\doteq-\sum_{m\in\mathcal{M}}&{\log(\mathbf{z}_{m}^\mathrm{GT}|\mathbf{z}_{m}^\mathrm{vis})},
\end{eqnarray}
where the loss is computed between the extracted features,
\begin{eqnarray}
\label{eqn:BEIT_2}
\{\mathbf{z}_{1}^\mathrm{vis},\ldots,\mathbf{z}_{m}^\mathrm{vis},\ldots,\mathbf{z}_{M}^\mathrm{vis}\}&={f^\mathrm{head}(f^\mathrm{vis}(\{\mathbf{t}_\mathrm{CLS},\mathbf{t}_1,..,\hat{\mathbf{t}}_m..,\mathbf{t}_M\}))},
\end{eqnarray}
and ground-truth guidance,
\begin{eqnarray}
\label{eqn:BEIT_2}
\{\mathbf{z}_{1}^\mathrm{GT},\ldots,\mathbf{z}_{m}^\mathrm{GT},\ldots,\mathbf{z}_{M}^\mathrm{GT}\}&={\mathrm{Tokenizer}(\{\mathbf{p}_1,\ldots,\mathbf{p}_m,\ldots,\mathbf{p}_M\})}.
\end{eqnarray}
In the above equations, $\mathcal{M}$ denotes the set of masked tokens, $\mathrm{Tokenizer}(\cdot)$ denotes the utilized vision tokenizer (\emph{e.g.}, d-VAE in BEIT) to extract visual features from the given patches, $f^\mathrm{head}(\cdot;\boldsymbol{\theta})$ represents the prediction head, and $\mathbf{t}_\mathrm{CLS}$ denotes the \textsf{CLS} token in the vision transformer, respectively. In this work, we also utilize a similar framework with BEIT but modify the prediction target with the guidance of multimodal knowledge.

\subsection{Multimodality-guided Visual Pre-training}
\begin{figure}[!t]
\centering
\includegraphics[width=12cm]{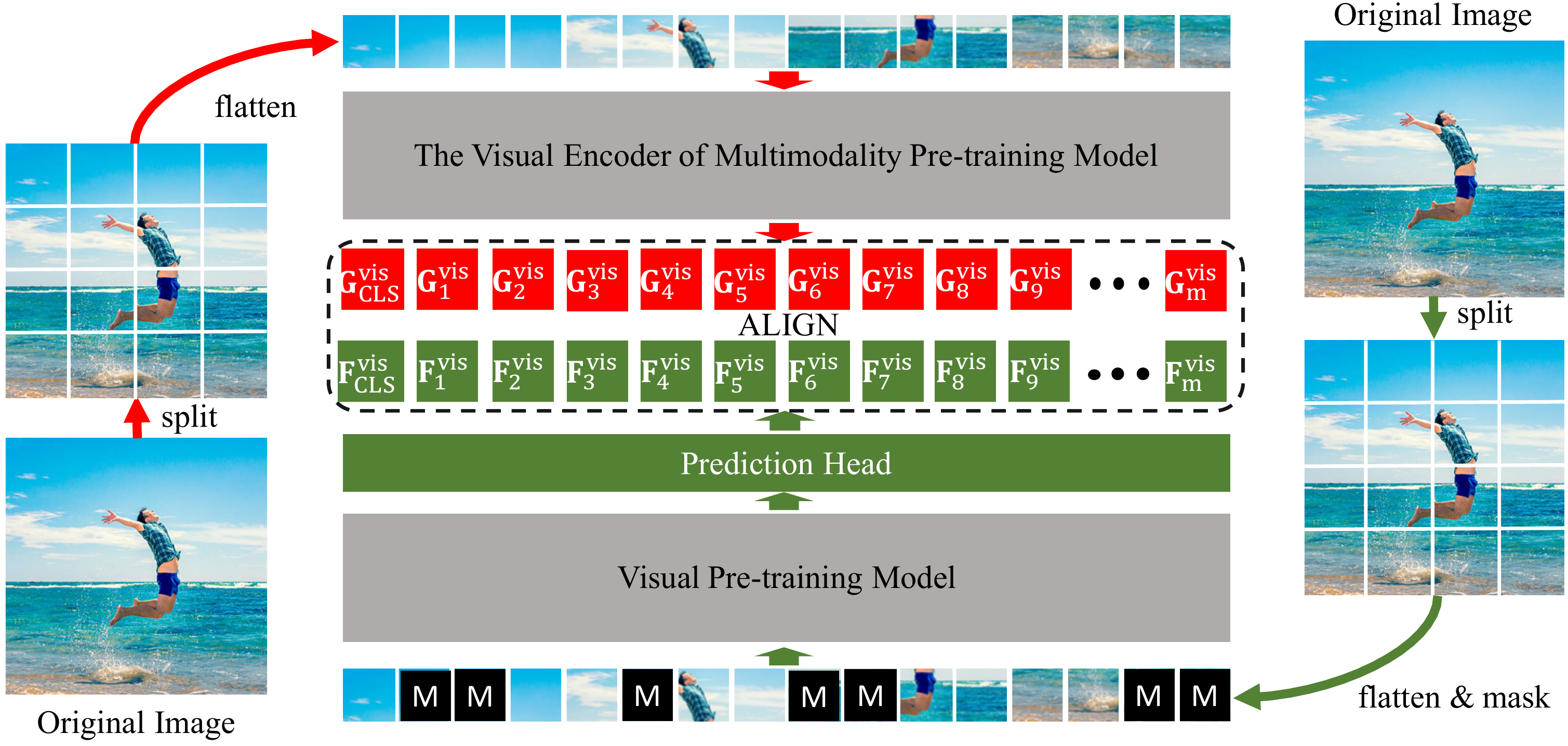}
\caption{Framework of the proposed MVP, where each \textsf{M} (\textsf{MASK}) denotes a masked token, and $\mathbf{F}_m^\mathrm{vis}$/$\mathbf{G}_m^\mathrm{vis}$ (or $\mathbf{F}_\mathrm{CLS}^\mathrm{vis}$/$\mathbf{G}_\mathrm{CLS}^\mathrm{vis}$) denotes the extracted features of each normal (or \textsf{CLS}) token. MVP is designed with the token-level multimodal information prediction pretext task to guide the pre-training of visual model.}
\label{fig:method}
\end{figure}
As described in Section~\ref{sec:intro}, the semantic discrimination of representation learned by previous MIM-based methods is relatively weak, for that they lack explicit semantics learning guidance. Target to address this problem, we require the tokenizer in MIM-based methods~\cite{BEIT,PeCo} to be aware of semantic information. To 
leverage weak supervision but not bias towards a specific semantic space, we decide to use a tokenizer pre-trained by multimodal data. Specially, we resort to a pre-trained multimodal model to extract semantic knowledge of each token, and then pre-train visual models with a multimodal knowledge prediction task. The proposed scheme is named Multimodality-guided Visual Pre-training (MVP), and the overview of MVP is shown in Figure~\ref{fig:method}. In the following, we will introduce the details of each module in MVP.

\subsubsection{Multimodal Semantics Extraction.} The goal of this module is to extract discriminative semantic knowledge but not bias towards a specific semantic space for the visual pre-training purpose. As shown in Figure~\ref{fig:introduction}, the language can better describe the semantics inside each image compared with the single annotated label. Motivated by this, our MVP utilizes a recent state-of-the-art multimodal pre-training model (CLIP, which has been pre-trained on 400 million image-text pairs) to extract multimodal semantic knowledge. Given the large-scale image-text pairs as ${\{(\mathbf{T}_n,\mathbf{J}_n)\}_{n=1}^{N'}}$, where $\mathbf{T}_n$ describes the semantic content of each image $\mathbf{J}_n$ (to be distinguished from $\mathbf{I}_n$), the optimization goal of CLIP can be simply formulated as:
 \begin{eqnarray}
\label{eqn:clip}
{\mathrm{Dist}(\mathbf{J}_n,\mathbf{T}_n)}&<&{\forall_{{n'\ne{n}}}\mathrm{Dis}(\mathbf{J}_n,\mathbf{T}_{n'})},
\end{eqnarray}
in which
 \begin{eqnarray}
{\mathrm{Dis}(\mathbf{J}_n,\mathbf{T}_{n'})}&=&{\langle {g^\mathrm{vis}}(\mathbf{J}_n),{g^\mathrm{lang}}(\mathbf{T}_{n'})\rangle},
\end{eqnarray}
where $\langle\cdot,\cdot\rangle$ denotes the cosine distance measurement. $g^\mathrm{vis}(\cdot)$ and $g^\mathrm{lang}(\cdot)$ represent the vision and language branch of the multimodal model (\textit{i.e.}, CLIP), respectively. Benefiting from Eqn (\ref{eqn:clip}), whether the visual feature extracted by the vision branch or the text feature extracted by the language branch is finally projected into a common multimodal space, and the semantics of this space is discriminative undoubtedly for it pre-training on the huge image-text pairs.

To integrate with the MIM framework, our MVP chooses the transformer architecture as vision branch of CLIP to extract multimodal semantics. Thus, the corresponding extracted multimodal knowledge of each token can be represented as:
 \begin{eqnarray}
\label{eqn:clip_vis}
{\{\mathbf{G}_\mathrm{CLS}^\mathrm{vis},\mathbf{G}_1^\mathrm{vis},\ldots,\mathbf{G}_m^\mathrm{vis},\ldots,\mathbf{G}_M^\mathrm{vis}\}}&=&{{g^\mathrm{vis}}(\{\mathbf{t}_\mathrm{CLS},\mathbf{t}_1,\ldots,\mathbf{t}_m,\ldots,\mathbf{t}_M\})},
\end{eqnarray}
where $\mathbf{G}_m^\mathrm{vis}$ denotes the feature of visual token $\mathbf{t}_m$, and $\mathbf{G}_\mathrm{CLS}^\mathrm{vis}$ represents the global feature extracted on the \textsf{CLS} token.

\subsubsection{Multimodal Information Prediction.} After obtaining the multimodal feature of each token, MVP further utilizes the designed multimodal information prediction pretext task to guide the pre-training of visual models. Same with BEIT, MVP firstly uses the Blockwise Masking scheme~\cite{BEIT} to mask percentages of tokens, and then inputs these masked tokens and remained unmasked tokens into the visual model to extract visual features. Furthermore, one extra prediction head $P^\mathrm{head}(\cdot)$ as BEIT is added to project these token-level visual features into the multimodal space. Therefore, the predicted multimodal information of each visual token can be formulated as:
\begin{eqnarray}
\label{eqn:MVP}
{\{\mathbf{F}_\mathrm{CLS}^\mathrm{vis},\mathbf{F}_1^\mathrm{vis},\ldots,\mathbf{F}_m^\mathrm{vis},\ldots,\mathbf{F}_M^\mathrm{vis}\}}&=&{f^\mathrm{head}}(f^\mathrm{vis}(\{\mathbf{t}_\mathrm{CLS},\mathbf{t}_1\ldots,\hat{\mathbf{t}}_m,\ldots,\mathbf{t}_M\})),
\end{eqnarray}
where $\mathbf{F}_m^\mathrm{vis}$ denotes the predicted multimodal feature of visual token $\hat{\mathbf{t}}_m$ according to the visual models, and $\mathbf{F}_\mathrm{CLS}^\mathrm{vis}$ represents the predicted global multimodal feature of the \textsf{CLS} token.

After obtaining the predicted multimodal feature of each token according to Eqn (\ref{eqn:MVP}) and the corresponding ground truth generated by Eqn (\ref{eqn:clip_vis}), MVP can guide the pre-training of visual models by achieving the token-level alignment of these features:
\begin{equation}\label{eq:Alignment}
    \mathcal{L}_\mathrm{MVP}\doteq -\frac{\langle{\mathbf{F}_\mathrm{CLS}^\mathrm{vis}},{\mathbf{G}_\mathrm{CLS}^\mathrm{vis}\rangle+\sum_{m=1}^M\langle{\mathbf{F}_{m}^\mathrm{vis}},{\mathbf{G}_{m}^\mathrm{vis}}\rangle}}{M+1},
\end{equation}

Driven by the alignment of Eqn (\ref{eq:Alignment}), MVP can guide the candidate visual model well to learn the common semantic knowledge of different modalities both in the global level (the alignment on the \textsf{CLS} token) and patch level (the alignment on each visual token). Thus, compared with recent self-supervised visual pre-training approaches, the pre-trained knowledge of visual models driven by MVP will contain more discriminative but relatively unbiased information. More evaluation and analysis of MVP are introduced in Section~\ref{sec:evaluation}.

\subsubsection{Implementation Details.} 
Following most MIM-based approaches~\cite{MAE,BEIT,PeCo}, we mainly utilize a series of ViT backbones~\cite{VIT} to evaluate the effectiveness of MVP. To achieve the token-level alignment as Eqn (\ref{eq:Alignment}), we directly utilize the vision branch with ViT-Base/16 backbone of CLIP, to extract multimodal knowledge. Notably, whether pre-training the visual model with ViT-Base/16 or ViT-Large/16 backbone, the backbone of vision branch in CLIP is always selected with ViT-Base/16. During pre-training, the parameters of vision branch in CLIP are frozen, and only of the parameters in candidate pre-trained visual model are tuned. For all variants of ViT in this paper, the image resolution of the input is set as $224\times224$, and Blockwise Masking scheme is employed to mask $75$ visual tokens as BEIT~\cite{BEIT}. Additionally, AdamW optimizer and a cosine decay learning rate scheduler are utilized. The initial learning rate and weight decay are set as $1.5$e-3 and $5$e-2, respectively. Same with most previous works~\cite{DINO,MaskFeat}, MVP is pre-trained on ImageNet-1K lasting for $300$ epochs. In the following of this paper, we denote ViT-Base/16 as ViT-B/16, and ViT-Large/16 as ViT-L/16 for short, respectively.

\subsection{Relationship to Prior Work}
Some concurrent works~\cite{Virtex,MCT} seem similar to our MVP, which also utilize language to guide the pre-training of visual models. However, there are still differences in motivations and implementations with ours. Virtex~\cite{Virtex} utilized a heavy-weight textual prediction head (vision transformer architecture) to process the visual feature extracted by the convolutional network, and then predict the corresponding caption of each image. Though Virtex has achieved competitive transfer performances on different visual downstream tasks, there still exists a problem that the pre-training capability of visual models heavily relies on the textual head. However, the image caption prediction task is very challenging for that each image can be described by different texts. Therefore, Virtex is hard to be pre-trained on super large-scale image-text datasets. Differently, our MVP simply designs the multimodal semantics prediction task on visual models with an additional light-weight prediction head (only one fully-connected layer), which guides the pre-training of visual models to learn the common semantic knowledge of different modalities (provided by CLIP).

Additionally, MaskFeat~\cite{MaskFeat} is the most similar work with ours currently. It also designs a feature prediction task based on the masked image modeling framework. However, MaskFeat only utilizes the visual feature (\emph{e.g.}, HOG feature) to supervise the pre-training of visual models, which is relatively weak from the view of semantics discrimination. Differently, the motivation of our MVP is resorting to the semantics guidance of different modalities. To achieve this, MVP utilizes a pre-trained multimodal model to replace the tokenizer in BEIT, and designs a corresponding feature prediction pretext task. Extensive experiments show our MVP enjoys lots of benefits on a wide range of downstream tasks compared with MaskFeat. More comparisons and analysis have been shown in Section~\ref{sec:evaluation}.

\section{Experimental Results}
\label{sec:evaluation}

\subsection{Datasets and Downstream Evaluation Setup} Same with most previous works~\cite{BEIT,MAE,PeCo}, MVP is mainly evaluated on image classification and semantic segmentation tasks. The details of our utilized datasets and experimental settings are introduced in the next.

\textbf{Datasets.} In this paper, MVP is mainly evaluated on the image classification task of ImageNet-1K, which contains about 130 million labeled images. Additionally, ADE20K~\cite{ADE20K} is a relatively challenging semantic segmentation dataset, and it contains $25$K images of 150 categories. In this paper, we also conduct evaluations on the semantic segmentation task of ADE20K. Compared with image classification, semantic segmentation is a dense vision task, which contains multiple instances inside each image. Therefore, the evaluation on semantic segmentation can better reflect the semantics capacity of pre-trained visual models.

\textbf{Image Classification Setup.}  While end-to-end fine-tuning the visual models pre-trained by MVP, we follow the most of hyper-parameter settings in the work~\cite{BEIT}. AdamW optimizer is utilized, and the weight decay is set as $5$e-$2$. The initial learning rate is set with $4$e-$3$ for ViT-B/16, and $1$e-$3$ for ViT-L/16, respectively. Additionally, a cosine decay learning rate scheduler is applied. We fine-tune the ViT-B/16 last for $100$ epochs with $20$ warm-up epochs, and $50$ epochs with $5$ warm-up epochs for ViT-L/16.

\textbf{Semantic Segmentation Setup.} While fine-tuning the pre-trained visual models on semantic segmentation task of ADE20K, we also follow the most of hyper-parameter settings in BEIT~\cite{BEIT}, in which the resolution of input image is set as $512\times512$, AdamW optimizer is applied and the initial learning rate is set as $3$e-$4$ for ViT-B/16, and $2$e-$5$ for ViT-L/16, respectively. The batch size is set as $16$, and the models are fine-tuned last for $160$K steps.

\subsection{Comparisons on Image Classification}

In this section, we firstly conduct comparisons with recent state-of-the-art visual pre-training approaches on ImageNet-1K with end-to-end fine-tuning mode.
As shown in Table~\ref{tab:SOTA1}, our MVP achieves consistently competitive results while pre-training different visual models. For example, MVP achieves $84.4\%$ Top-1 accuracy of ViT-B/16 on ImageNet-1K, which outperforms the baseline (BEIT) with $1.2\%$. As for the ViT/L-16 backbone, MVP also surpasses BEIT with $1.1\%$ on Top-1 accuracy. Additionally, compared with MaskFeat, which also utilizes a feature prediction scheme similar to MVP, our work still shows better transfer performances while fine-tuning on ImageNet-1K,~\emph{e.g}, $0.4\%$ improvement on ViT/B-16 and $0.6\%$ improvement on ViT/L-16, respectively.

MVP is also evaluated on the linear probing test of ImageNet-1K, and it achieves $75.4\%$ Top-1 accuracy, which significantly outperforms the current MIM-based methods (\emph{e.g.}, BEIT reports a $37.6\%$ accuracy and MAE achieves a $67.8\%$ accuracy). The above consistent improvements clearly demonstrate the superiority of our multimodality-guided visual pre-training scheme. More evaluations and analysis can be seen in Section~\ref{sec:analysis}. We admit that there is still a weak performance gap in the linear probing task of MVP compared with previous self-supervised learning approaches (\emph{e.g.}, DINO reports a $78.2\%$ accuracy), which could own to the MIM framework and will be left for future study.

\begin{table}[!t]
\centering
\setlength{\abovecaptionskip}{10pt}%
\caption{Comparison to the state-of-the-arts on ImageNet-1K. All entries are firstly pre-trained on ImageNet-1K, and then further fine-tuned with end-to-end mode. The image resolution of the input for all entries is set as $224\times224$.}
\begin{tabular}{p{3cm}<{\centering}|p{2cm}<{\centering}|c<{\centering}|p{1.8cm}<{\centering}}
\thickhline
\textbf{Method} & \textbf{Model} & \textbf{Pre-training Epochs} & \textbf{Top-1 ($\%$)} \\
\thickhline
\noalign{\smallskip}
DINO~\cite{DINO}  & ViT-B/16 & $300$ & $82.8$  \\
BEIT~\cite{BEIT}  & ViT-B/16 & $800$ & $83.2$  \\
MAE~\cite{MAE} & ViT-B/16 & $1600$ & $83.6$\\
PeCo~\cite{PeCo} & ViT-B/16 & $300$ & $84.1$\\
MaskFeat~\cite{MaskFeat} & ViT-B/16 & $1600$ & $84.0$\\
\textbf{MVP (ours)} & \textbf{ViT-B/16} & \textbf{300} & \textbf{84.4}\\
\noalign{\smallskip}
\hline
\noalign{\smallskip}
BEIT~\cite{BEIT}  & ViT-L/16 & $800$ & $85.2$  \\
MAE~\cite{MAE}  & ViT-L/16 & $1600$ & $85.9$  \\
MaskFeat~\cite{MaskFeat}  & ViT-L/16 & $1600$ & $85.7$  \\
\textbf{MVP (ours)}  & \textbf{ViT-L/16} & \textbf{300} & \textbf{86.3}  \\
\thickhline
\end{tabular}
\label{tab:SOTA1}
\end{table}

\subsection{Comparisons on Semantic Segmentation}

Different from the image classification where the single object is present in each image, semantic segmentation is a more challenging task and each image contains multiple instances. Given that the text can fully describe the presented instances and their relationships inside each image, our MVP driven by multimodal information should achieve much better transfer performance on semantic segmentation datasets. 

To evaluate this, we conduct extensive comparisons with recent state-of-the-arts on ADE20K. As shown in Table~\ref{tab:SOTA2}, our MVP shows significant advantages on this task. For example, as for the recent MIM-based methods with different visual feature or pixel information reconstruction pretext tasks, they achieve the nearly same transfer performance,~\emph{e.g.}, $45.6\%$ mIoU of BEIT and $48.1\%$ mIoU of MAE on ViT/B-16, respectively. Differently, with the guidance of multimodal semantic knowledge, our MVP heavily pushes forward the transfer results,~\emph{e.g.}, $3.6\%$ improvement compared with the previous best reported performance of ViT-B/16.

It is admitted that the above significant improvements of our MVP could own to the super large-scale multimodal dataset while pre-training CLIP. To valid this, we also conduct comparisons with the BEIT model pre-trained on ImageNet-21K, which contains about $21$K classes. As shown in Table~\ref{tab:ADE2}, MVP still shows consistent advantages in transferring pre-trained knowledge. Notably, there are lots of coarse aligned image-text pairs on websites, and they are nearly free to be available. Therefore, the requirement of MVP with a pre-trained multimodal model is not a hard constrain for the visual pre-training community.
\begin{table}[!t]
\centering
\setlength{\abovecaptionskip}{10pt}%
\caption{Comparison to the state-of-the-arts on ADE20K. All models are pre-trained on ImageNet-1K and fine-tuned on ADE20K. The image resolution of the input for all entries is set as $512\times512$.}
\begin{tabular}{p{3cm}<{\centering}|p{2cm}<{\centering}|c<{\centering}|p{1.8cm}<{\centering}}
\thickhline
\textbf{Method} & \textbf{Model} & \textbf{Pre-training Epochs} & \textbf{mIoU ($\%$)} \\
\thickhline
\noalign{\smallskip}
DINO~\cite{DINO}  & ViT-B/16 & 300 & $44.1$  \\
BEIT~\cite{BEIT}  & ViT-B/16 & $800$ & $45.6$  \\
MAE~\cite{MAE} & ViT-B/16 & $1600$ & $48.1$\\
CAE~\cite{CAE} & ViT-B/16 & $800$ & $48.8$\\
PeCo~\cite{PeCo} & ViT-B/16 & $300$ & $46.7$\\
\textbf{MVP (ours)} & \textbf{ViT-B/16} & \textbf{300} & \textbf{52.4}\\
\thickhline
\end{tabular}
\label{tab:SOTA2}
\end{table}

\begin{table}[!t]
\centering
\setlength{\abovecaptionskip}{10pt}%
\caption{Comparisons of BEIT and our MVP pre-trained on different datasets. All evaluations are conducted on ADE20K, and the image resolution of the input for all models is set as $512\times512$. BEIT$^{*}$ represents that we reproduce its result on ADE20K with the officially released model.}
\begin{tabular}{p{3cm}<{\centering}|p{2cm}<{\centering}|c<{\centering}|p{1.8cm}<{\centering}}
\thickhline
\textbf{Method} & \textbf{Model} & \textbf{Pre-training Dataset} & \textbf{mIoU ($\%$)} \\
\thickhline
\noalign{\smallskip}
BEIT$^{*}$~\cite{BEIT}  & ViT-B/16 & ImageNet-21K & $46.3$  \\
\textbf{MVP (ours)} & \textbf{ViT-B/16} & \textbf{ImageNet-1K} & \textbf{52.4}\\
\noalign{\smallskip}
\hline
\noalign{\smallskip}
BEIT$^{*}$~\cite{BEIT}  & ViT-L/16 & ImageNet-21K & $51.8$  \\
\textbf{MVP (ours)}  & \textbf{ViT-L/16} & \textbf{ImageNet-1K} & \textbf{54.3}  \\
\thickhline
\end{tabular}
\label{tab:ADE2}
\end{table}

\subsection{Ablation Study}
\label{sec:analysis}
\textbf{The effect of different pre-training epochs.} In this section, we first verify the effect of different pre-training epochs on MVP. As shown in Table~\ref{tab:AB_epoch}, with the pre-training epochs increasing, the transfer performance of MVP is gradually improved,~\emph{e.g.}, while enlarging the pre-training epochs from $100$ to $300$, there is a $0.5\%$ improvement on Top-1 accuracy of ImageNet-1K and $0.4\%$ improvement on mIoU metric of ADE20K, respectively. 
\begin{table}[!t]
\centering
\setlength{\abovecaptionskip}{10pt}%
\caption{The effect of different pre-training epochs on MVP.}
\begin{tabular}{p{2cm}<{\centering}|p{1.7cm}<{\centering}|p{3.5cm}<{\centering}|p{3.0cm}<{\centering}}
\thickhline
 \textbf{Model} & \textbf{Epochs} &  \textbf{ImageNet-1K(Top-1)} & \textbf{ADE20K(mIoU)} \\
\thickhline
\noalign{\smallskip}
ViT-B/16 & 100 & $83.9$ & $52.0$  \\
ViT-B/16 & 300 & $84.4$ & $52.4$  \\
\thickhline
\end{tabular}
\label{tab:AB_epoch}
\end{table}

\noindent\textbf{The effect of guidance with different knowledge.} There is another question is whether the improvement of MVP is taken by the designed feature prediction pretext task or the guidance of multimodal semantic knowledge. One indirect evidence is that our MVP achieves much better transfer performances on downstream tasks compared with MaskFeat, a recent state-of-the-art with the pretext task of reconstructing the HOG feature. However, the discrimination of HOG feature is relatively weak. To better evaluate it, we resort to the guidance of a recent self-supervised visual pre-training method, DINO~\cite{DINO}. Similar to MVP, we firstly utilize the ViT-B/16 backbone pre-trained by DINO to extract its visual feature of each token, and then these features are regarded as the prediction target to guide the visual models pre-training. The comparisons are shown in Table~\ref{tab:AB_guidance}. Generally, the model with guidance of a pre-trained multimodal model can achieve much better results compared with the model driven by visual pre-training model, especially evaluated on dense vision downstream task,~\emph{i.e.}, $5.4\%$ improvement on ADE20K. The above significant improvement further evaluates the superiority of our multimodality-guided visual pre-training scheme.
\begin{table}[!t]
\centering
\setlength{\abovecaptionskip}{10pt}%
\caption{The effect of utilizing different guidance on MVP.}
\begin{tabular}{p{1.7cm}<{\centering}|p{1.7cm}<{\centering}|p{1.3cm}<{\centering}|p{3.5cm}<{\centering}|p{3.0cm}<{\centering}}
\thickhline
 \textbf{Guidance} & \textbf{Model} & \textbf{Epochs} &  \textbf{ImageNet-1K(Top-1)} & \textbf{ADE20K(mIoU)} \\
\thickhline
\noalign{\smallskip}
DINO & ViT-B/16 & 300 & $83.6$ & $47.0$  \\
CLIP & ViT-B/16 & 300 & $84.4$ & $52.4$  \\
\thickhline
\end{tabular}
\label{tab:AB_guidance}
\end{table}

\begin{figure}[H]
\centering
\includegraphics[width=12cm]{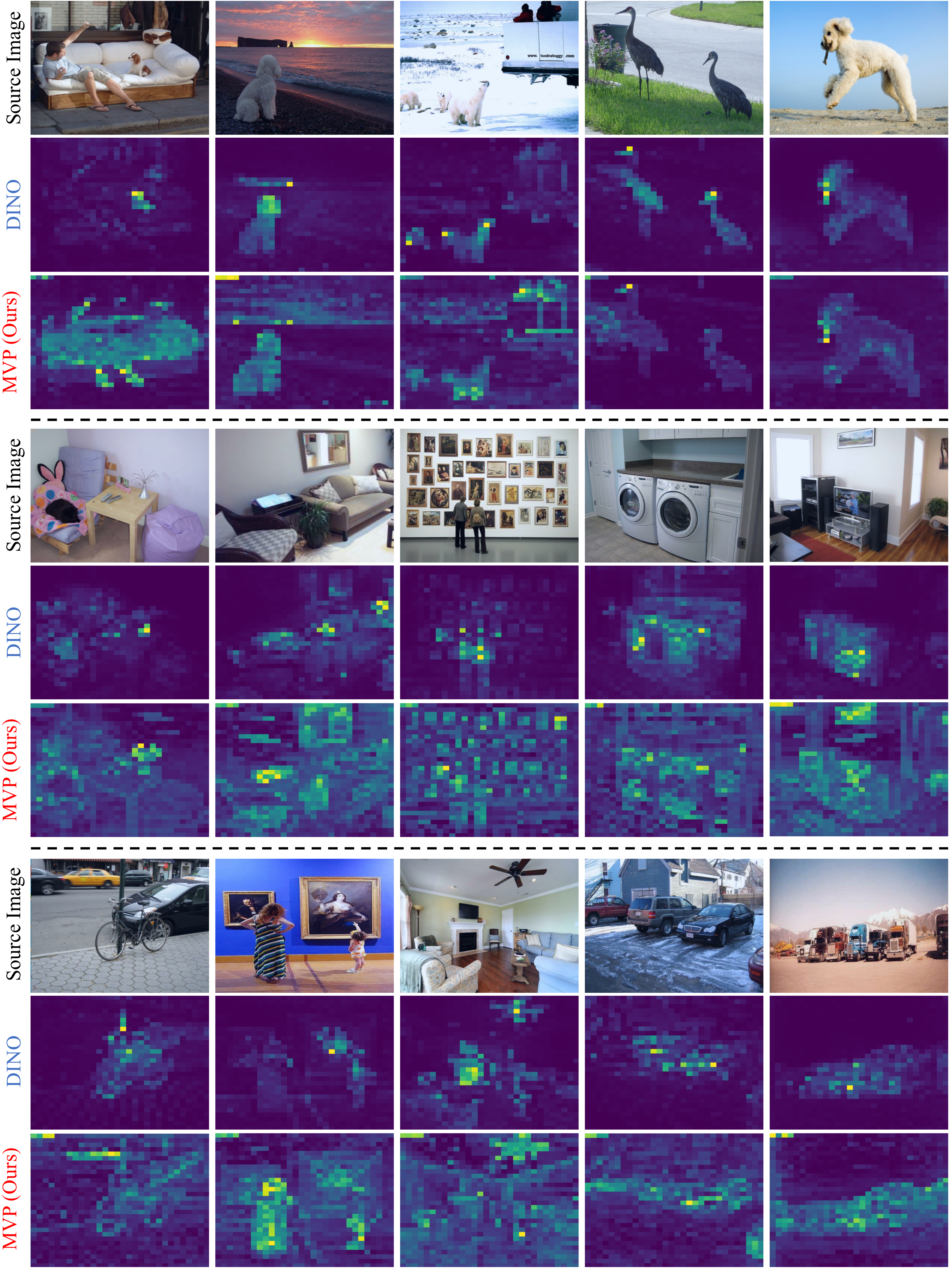}
\caption{Visualization of representation learned by different visual pre-training methods. The presented feature map is generated by averaging their attention maps of all heads in ViT-B/16 with \textsf{CLS} token as the query. Images in the first row are from ImageNet-1K, and images of the fourth and seventh row are from ADE20K. Generally, the previous self-supervised visual pre-training method (DINO) only attends on limited regions. Differently, with the guidance of multimodal semantics, MVP can handle the complex image and better describe the overall scene and instances inside each image.}
\label{fig:visualization}
\end{figure}

\noindent\textbf{Analysis of representation learning in MVP.} Though MVP achieves excellent transfer performances on different downstream tasks, there remains a question of whether it learns dense visual features as the corresponding language describes. To evaluate this, we conduct visualization analysis on the representation of MVP. For that there is no explicit constrain on \textsf{CLS} token of BEIT to extract the global feature, we cannot conduct comparisons on the learned global representation of BEIT and our MVP. To analyze the character of representation in MVP, we utilize DINO, a recent state-of-the-art self-supervised learning method, to compare. For both MVP and DINO, we use \textsf{CLS} token as the query to look for the corresponding response map. We average their attention maps of all heads in ViT-B/16 backbone and then present the results in Figure~\ref{fig:visualization}.

Generally, the previous self-supervised learning method (DINO) only focuses on the specified foreground (\emph{e.g.}, the dog and the polar bear), but ignores other useful information. Differently, MVP can better describe the overall scene and instances inside each image,~\emph{e,g.}, ``a man and his dog play together in the sofa'', ``the dog is seeing the sunset'', and \textit{etc}. Moreover, as for the complex images in ADE20K,~\emph{e.g}, ``there are two chairs and one table in this room" and ``peoples are looking at the multiple murals on the wall'', MVP seems to have the ability to extract the whole of its content inside each image from the view of attention map, which could be the reason of that MVP performs much better than previous visual pre-training works on the semantic segmentation task.

As discussed above, it is concluded that MVP can better learn the multi-grained dense semantic knowledge inside each image. This phenomenon further reflects the superiority of visual pre-training driven by multimodal semantic knowledge. In the future, we will evaluate our MVP on more dense vision downstream tasks.

\subsection{Discussions and Future Perspectives}

First, we shall recognize that MVP relies on the representation ability brought by multimodal pre-training. Therefore, the comparison between MVP and pure vision pre-training algorithms is not completely fair. However, this paper hopes to deliver the message that pure vision pre-training, especially the recent MIM-based approaches~\cite{BEIT,MAE}, suffers the limitations of learning semantic information -- this seems not to be solved by simply using larger datasets (\textit{e.g.}, ImageNet-21K, used by BEIT). We advocate for using of multimodal information towards a potential breakthrough.

Second, we notice that MVP, like BEIT, is built upon a pre-trained tokenizer and thus incurs extra training costs. On the other hand, provided a multimodal pre-trained model, MVP enjoys a higher training efficiency in the pure vision domain. Going further along this direction may enlighten the community to establish multimodal pre-training as an upstream task of single-modal pre-training. In the future, it will be interesting to extend the idea to more languages~\cite{M3p} and even more data modalities~\cite{data2vec}, observing their contribution to visual representation learning.

\section{Conclusion}
In this paper, we present Multimodality-guided Visual Pre-training (MVP), the first work to introduce guidance from other modalities on masked image modeling. By replacing the tokenizer with the vision branch of CLIP on BEIT and simply modifying the prediction task, MVP can better learn the multimodal semantic knowledge inside each image. 
Extensive experiments on a wide range of visual downstream tasks have clearly shown the effectiveness of MVP on pre-training visual models. 
Importantly, this work points a new direction for visual pre-training with other modalities. In the future, the effective visual pre-training schemes with more data modalities guidance will be designed.

\bibliographystyle{splncs04}
\bibliography{egbib}

\begin{thebibliography}{10}
\providecommand{\url}[1]{\texttt{#1}}
\providecommand{\urlprefix}{URL }
\providecommand{\doi}[1]{https://doi.org/#1}

\bibitem{data2vec}
Baevski, A., Hsu, W.N., Xu, Q., Babu, A., Gu, J., Auli, M.: Data2vec: A general
  framework for self-supervised learning in speech, vision and language. arXiv
  preprint arXiv:2202.03555  (2022)

\bibitem{BEIT}
Bao, H., Dong, L., Wei, F.: Beit: Bert pre-training of image transformers.
  arXiv preprint arXiv:2106.08254  (2021)

\bibitem{DINO}
Caron, M., Touvron, H., Misra, I., J{\'e}gou, H., Mairal, J., Bojanowski, P.,
  Joulin, A.: Emerging properties in self-supervised vision transformers. In:
  Proceedings of the IEEE/CVF International Conference on Computer Vision. pp.
  9650--9660 (2021)

\bibitem{chen2020simple}
Chen, T., Kornblith, S., Norouzi, M., Hinton, G.: A simple framework for
  contrastive learning of visual representations. arXiv preprint
  arXiv:2002.05709  (2020)

\bibitem{CAE}
Chen, X., Ding, M., Wang, X., Xin, Y., Mo, S., Wang, Y., Han, S., Luo, P.,
  Zeng, G., Wang, J.: Context autoencoder for self-supervised representation
  learning. arXiv preprint arXiv:2202.03026  (2022)

\bibitem{chen2020improved}
Chen, X., Fan, H., Girshick, R., He, K.: Improved baselines with momentum
  contrastive learning. arXiv preprint arXiv:2003.04297  (2020)

\bibitem{MOCOV3}
Chen, X., Xie, S., He, K.: An empirical study of training self-supervised
  vision transformers. arXiv preprint arXiv:2104.02057  (2021)

\bibitem{Uniter}
Chen, Y.C., Li, L., Yu, L., El~Kholy, A., Ahmed, F., Gan, Z., Cheng, Y., Liu,
  J.: Uniter: Universal image-text representation learning. In: European
  conference on computer vision. pp. 104--120. Springer (2020)

\bibitem{deng2009imagenet}
Deng, J., Dong, W., Socher, R., Li, L.J., Li, K., Fei-Fei, L.: Imagenet: A
  large-scale hierarchical image database. In: 2009 IEEE conference on computer
  vision and pattern recognition. pp. 248--255. Ieee (2009)

\bibitem{Virtex}
Desai, K., Johnson, J.: Virtex: Learning visual representations from textual
  annotations. In: Proceedings of the IEEE/CVF Conference on Computer Vision
  and Pattern Recognition. pp. 11162--11173 (2021)

\bibitem{devlin2018bert}
Devlin, J., Chang, M.W., Lee, K., Toutanova, K.: Bert: Pre-training of deep
  bidirectional transformers for language understanding. arXiv preprint
  arXiv:1810.04805  (2018)

\bibitem{PeCo}
Dong, X., Bao, J., Zhang, T., Chen, D., Zhang, W., Yuan, L., Chen, D., Wen, F.,
  Yu, N.: Peco: Perceptual codebook for bert pre-training of vision
  transformers. arXiv preprint arXiv:2111.12710  (2021)

\bibitem{dosovitskiy2020image}
Dosovitskiy, A., Beyer, L., Kolesnikov, A., Weissenborn, D., Zhai, X.,
  Unterthiner, T., Dehghani, M., Minderer, M., Heigold, G., Gelly, S., et~al.:
  An image is worth 16x16 words: Transformers for image recognition at scale.
  arXiv preprint arXiv:2010.11929  (2020)

\bibitem{VIT}
Dosovitskiy, A., Beyer, L., Kolesnikov, A., Weissenborn, D., Zhai, X.,
  Unterthiner, T., Dehghani, M., Minderer, M., Heigold, G., Gelly, S., et~al.:
  An image is worth 16x16 words: Transformers for image recognition at scale.
  arXiv preprint arXiv:2010.11929  (2020)

\bibitem{grill2020bootstrap}
Grill, J.B., Strub, F., Altch{\'e}, F., Tallec, C., Richemond, P.H.,
  Buchatskaya, E., Doersch, C., Pires, B.A., Guo, Z.D., Azar, M.G., et~al.:
  Bootstrap your own latent: A new approach to self-supervised learning. arXiv
  preprint arXiv:2006.07733  (2020)

\bibitem{MAE}
He, K., Chen, X., Xie, S., Li, Y., Doll{\'a}r, P., Girshick, R.: Masked
  autoencoders are scalable vision learners. arXiv preprint arXiv:2111.06377
  (2021)

\bibitem{he2020momentum}
He, K., Fan, H., Wu, Y., Xie, S., Girshick, R.: Momentum contrast for
  unsupervised visual representation learning. In: Proceedings of the IEEE/CVF
  Conference on Computer Vision and Pattern Recognition. pp. 9729--9738 (2020)

\bibitem{he2019rethinking}
He, K., Girshick, R., Doll{\'a}r, P.: Rethinking imagenet pre-training. In:
  Proceedings of the IEEE/CVF International Conference on Computer Vision. pp.
  4918--4927 (2019)

\bibitem{Align}
Jia, C., Yang, Y., Xia, Y., Chen, Y.T., Parekh, Z., Pham, H., Le, Q., Sung,
  Y.H., Li, Z., Duerig, T.: Scaling up visual and vision-language
  representation learning with noisy text supervision. In: International
  Conference on Machine Learning. pp. 4904--4916. PMLR (2021)

\bibitem{lan2019albert}
Lan, Z., Chen, M., Goodman, S., Gimpel, K., Sharma, P., Soricut, R.: Albert: A
  lite bert for self-supervised learning of language representations. arXiv
  preprint arXiv:1909.11942  (2019)

\bibitem{Visualbert}
Li, L.H., Yatskar, M., Yin, D., Hsieh, C.J., Chang, K.W.: Visualbert: A simple
  and performant baseline for vision and language. arXiv preprint
  arXiv:1908.03557  (2019)

\bibitem{Oscar}
Li, X., Yin, X., Li, C., Zhang, P., Hu, X., Zhang, L., Wang, L., Hu, H., Dong,
  L., Wei, F., et~al.: Oscar: Object-semantics aligned pre-training for
  vision-language tasks. In: European Conference on Computer Vision. pp.
  121--137. Springer (2020)

\bibitem{M3p}
Ni, M., Huang, H., Su, L., Cui, E., Bharti, T., Wang, L., Zhang, D., Duan, N.:
  M3p: Learning universal representations via multitask multilingual multimodal
  pre-training. In: Proceedings of the IEEE/CVF Conference on Computer Vision
  and Pattern Recognition. pp. 3977--3986 (2021)

\bibitem{noroozi2016unsupervised}
Noroozi, M., Favaro, P.: Unsupervised learning of visual representations by
  solving jigsaw puzzles. In: European Conference on Computer Vision. pp.
  69--84. Springer (2016)

\bibitem{pathak2016context}
Pathak, D., Krahenbuhl, P., Donahue, J., Darrell, T., Efros, A.A.: Context
  encoders: Feature learning by inpainting. In: Proceedings of the IEEE
  conference on computer vision and pattern recognition. pp. 2536--2544 (2016)

\bibitem{2021Learning}
Radford, A., Kim, J.W., Hallacy, C., Ramesh, A., Goh, G., Agarwal, S., Sastry,
  G., Askell, A., Mishkin, P., Clark, J.: Learning transferable visual models
  from natural language supervision  (2021)

\bibitem{cc3m}
Sharma, P., Ding, N., Goodman, S., Soricut, R.: Conceptual captions: A cleaned,
  hypernymed, image alt-text dataset for automatic image captioning. In:
  Proceedings of the 56th Annual Meeting of the Association for Computational
  Linguistics (Volume 1: Long Papers). pp. 2556--2565 (2018)

\bibitem{tian2019contrastive}
Tian, Y., Krishnan, D., Isola, P.: Contrastive multiview coding. arXiv preprint
  arXiv:1906.05849  (2019)

\bibitem{tian2020makes}
Tian, Y., Sun, C., Poole, B., Krishnan, D., Schmid, C., Isola, P.: What makes
  for good views for contrastive learning. arXiv preprint arXiv:2005.10243
  (2020)

\bibitem{touvron2021training}
Touvron, H., Cord, M., Douze, M., Massa, F., Sablayrolles, A., J{\'e}gou, H.:
  Training data-efficient image transformers \& distillation through attention.
  In: International Conference on Machine Learning. pp. 10347--10357. PMLR
  (2021)

\bibitem{DEIT}
Touvron, H., Cord, M., Douze, M., Massa, F., Sablayrolles, A., J{\'e}gou, H.:
  Training data-efficient image transformers \& distillation through attention.
  In: International Conference on Machine Learning. pp. 10347--10357. PMLR
  (2021)

\bibitem{VQVAE}
Van Den~Oord, A., Vinyals, O., et~al.: Neural discrete representation learning.
  Advances in neural information processing systems  \textbf{30} (2017)

\bibitem{vincent2008extracting}
Vincent, P., Larochelle, H., Bengio, Y., Manzagol, P.A.: Extracting and
  composing robust features with denoising autoencoders. In: Proceedings of the
  25th international conference on Machine learning. pp. 1096--1103 (2008)

\bibitem{MaskFeat}
Wei, C., Fan, H., Xie, S., Wu, C.Y., Yuille, A., Feichtenhofer, C.: Masked
  feature prediction for self-supervised visual pre-training. arXiv preprint
  arXiv:2112.09133  (2021)

\bibitem{wei2019iterative}
Wei, C., Xie, L., Ren, X., Xia, Y., Su, C., Liu, J., Tian, Q., Yuille, A.L.:
  Iterative reorganization with weak spatial constraints: Solving arbitrary
  jigsaw puzzles for unsupervised representation learning. In: Proceedings of
  the IEEE Conference on Computer Vision and Pattern Recognition. pp.
  1910--1919 (2019)

\bibitem{wei2021exploring}
Wei, L., Xie, L., Zhou, W., Li, H., Tian, Q.: Exploring the diversity and
  invariance in yourself for visual pre-training task. arXiv preprint
  arXiv:2106.00537  (2021)

\bibitem{SimMIM}
Xie, Z., Zhang, Z., Cao, Y., Lin, Y., Bao, J., Yao, Z., Dai, Q., Hu, H.:
  Simmim: A simple framework for masked image modeling. arXiv preprint
  arXiv:2111.09886  (2021)

\bibitem{MCT}
Yuan, X., Lin, Z., Kuen, J., Zhang, J., Wang, Y., Maire, M., Kale, A., Faieta,
  B.: Multimodal contrastive training for visual representation learning. In:
  Proceedings of the IEEE/CVF Conference on Computer Vision and Pattern
  Recognition. pp. 6995--7004 (2021)

\bibitem{dvae}
Zhang, M., Jiang, S., Cui, Z., Garnett, R., Chen, Y.: D-vae: A variational
  autoencoder for directed acyclic graphs. Advances in Neural Information
  Processing Systems  \textbf{32} (2019)

\bibitem{ADE20K}
Zhou, B., Zhao, H., Puig, X., Xiao, T., Fidler, S., Barriuso, A., Torralba, A.:
  Semantic understanding of scenes through the ade20k dataset. International
  Journal of Computer Vision  \textbf{127}(3),  302--321 (2019)

\bibitem{ibot}
Zhou, J., Wei, C., Wang, H., Shen, W., Xie, C., Yuille, A., Kong, T.: ibot:
  Image bert pre-training with online tokenizer. arXiv preprint
  arXiv:2111.07832  (2021)

\end{thebibliography}
\end{document}